\documentclass[11pt,a4paper]{article}
\usepackage[nohyperref]{naaclhlt2019}
\usepackage{times}
\usepackage{latexsym}
\usepackage{booktabs}
\usepackage{amsmath}
\usepackage{graphicx}
\usepackage{multirow}
\usepackage{multicol}
\usepackage{qtree}

\aclfinalcopy 

\def\aclpaperid{1131} 

\title{Addressing Word-order Divergence in Multilingual Neural Machine Translation for extremely Low Resource Languages}

\author{Rudra Murthy V$^{\dagger}$, Anoop Kunchukuttan$^{\ddagger}$, Pushpak Bhattacharyya$^{\dagger}$ \\
  $^{\dagger}$ Center for Indian Language Technology (CFILT) \\
  Department of Computer Science and Engineering \\
IIT Bombay, India. \\
$^{\ddagger}$Microsoft AI \& Research, Hyderabad, India.\\
{\tt \{rudra,pb\}@cse.iitb.ac.in, ankunchu@microsoft.com} 
}

\date{}

\begin{document}
\maketitle
\begin{abstract}
Transfer learning approaches for Neural Machine Translation (NMT) trains a NMT model on an assisting language-target language pair (parent model) which is later fine-tuned for the source language-target language pair of interest (child model), with the target language being the same. In many cases, the assisting language has a different word order from the source language. We show that divergent word order adversely limits the benefits from transfer learning when little to no parallel corpus between the source and target language is available. To bridge this divergence, we propose to pre-order the assisting language sentences to match the word order of the source language and train the parent model. Our experiments on many language pairs show that bridging the word order gap leads to major improvements in the translation quality in extremely low-resource scenarios.
\end{abstract}

\section{Introduction}

Transfer learning for multilingual Neural Machine Translation (NMT) \cite{D16-1163,Y17-1038,nguyen2017transfer} attempts to improve the NMT performance on the \textit{source} to \textit{target} language pair (child task) using an \textit{assisting source} language (assisting to target language translation is the parent task). Here, the parent model is trained on the assisting and target language parallel corpus and the trained weights are used to initialize the child model. If source-target language pair parallel corpus is available, the child model can further be fine-tuned. The weight initialization reduces the requirement on the training data for the source-target language pair by transferring knowledge from the parent task, thereby improving the performance on the child task. 

However, the divergence between the source and the assisting language can adversely impact the benefits obtained from transfer learning. Multiple studies have shown that transfer learning works best when the languages are related \cite{D16-1163,nguyen2017transfer,Y17-1038}. \citet{D16-1163} studied the influence of language divergence between languages chosen for training the parent and the child model, and showed that choosing similar languages for training the parent and the child model leads to better improvements from transfer learning. 

Several studies have tried to address the \textit{lexical divergence} between the source and the target languages either by using Byte Pair Encoding (BPE) as basic input representation units \cite{nguyen2017transfer} or character-level NMT system \cite{leeCharacterNMTTACL} or bilingual embeddings \cite{universalNMTGu}. However, the effect of \textit{word order divergence} and its mitigation has not been explored. In a practical setting, it is not uncommon to have source and assisting languages with different word order. For instance, it is possible to find parallel corpora between English (SVO word order) and some Indian (SOV word order) languages, but very little parallel corpora between Indian languages. Hence, it is natural to use English as an assisting language for inter-Indian language translation.

To address the word order divergence, we propose to pre-order the assisting language sentences (SVO) to match the word order of the source language (SOV). We consider an extremely resource-constrained scenario, where there is no parallel corpus for the child task. From our experiments, we show that there is a significant increase in the translation accuracy for the unseen source-target language pair. 

\section{Related Work}

To the best of our knowledge, no work has addressed word order divergence in transfer learning for multilingual NMT. However, some work exists for other NLP tasks in a multilingual setting. For Named Entity Recognition (NER), \citet{2018arXiv180809861X} use a self-attention layer after the Bi-LSTM layer to address word-order divergence for Named Entity Recognition (NER) task. The approach does not show any significant improvements, possibly because the divergence has to be addressed before/during construction of the contextual embeddings in the Bi-LSTM layer. \citet{joty2017cross} use adversarial training for cross-lingual question-question similarity ranking. The adversarial training tries to force the sentence representation generated by the encoder of similar sentences from different input languages to have similar representations.

Pre-ordering the source language sentences to match the target language word order has been found useful in addressing word-order divergence for Phrase-Based SMT \cite{collinsClause,ananthaGeneric,navratil2012comparison,DBLP:conf/icon-nlp/ChatterjeeKB14}. For NMT, \citet{pontiIsomorphic} and \citet{chenReorderJapan} have explored pre-ordering. \citet{pontiIsomorphic} demonstrated that by reducing the syntactic divergence between the source and the target languages, consistent improvements in NMT performance can be obtained. On the contrary, \citet{chenReorderJapan} reported drop in NMT performance due to pre-ordering. Note that these works address source-target divergence, not divergence between source languages in multilingual NMT scenario. 

\section{Proposed Solution}

Consider the task of translating for an extremely low-resource language pair. The parallel corpus between the two languages, if available may be too small to train an NMT model. Similar to \citet{D16-1163}, we use transfer learning to overcome data sparsity between the source and the target languages. We choose \textit{English} as the assisting language in all our experiments. In our resource-scarce scenario, we have no parallel corpus for training the child model. Hence, at test time, the source language sentence is translated using the parent model after performing a word-by-word translation from source to the assisting language using a bilingual dictionary. 

\begin{table}
    \centering
    \begin{tabular}{c|c}
    \toprule
    \textbf{Before Reordering} & \textbf{After Reordering}\\
    \midrule
    \Tree [.S [.NP$_0$ ] [.VP [.V ] [.NP$_1$ ] ] ]  &  \Tree [.S [.NP$_0$ ] [.VP [.NP$_1$ ] [.V ] ] ] \\
    \midrule
    {\scriptsize
    \begin{tabular}{c}
        \qtreepadding 2pt
        \Tree [.S [.NP [.NNP Anurag ] ] [.VP [.MD will ] [.VP [.VB meet ] [.NP [.NNP Thakur ] ] ] ] ]    
        \\
    \end{tabular} }
      & {\scriptsize
    \begin{tabular}{c}
        \qtreepadding 2pt
        \Tree [.S [.NP [.NNP Anurag ] ] [.VP [.NP [.NNP Thakur ] ] [.VP [.MD will ] [.VP [.VB meet ] ] ] ] ]
        \\
    \end{tabular} } \\
    \bottomrule
    \end{tabular}
    \caption{Example showing transitive verb before and after reordering (Adapted from  \citet{DBLP:conf/icon-nlp/ChatterjeeKB14})}
    \label{tab:reOrderingRule}
\end{table}

Since the source language and the assisting language (English) have different word order, we hypothesize that it leads to inconsistencies in the contextual representations generated by the encoder for the two languages. Specifically, given an English sentence (SVO word order) and its translation in the source language (SOV word order), the encoder representations for words in the two sentences will be different due to different contexts of synonymous words. This could lead to the attention and the decoder layers generating different translations from the same (parallel) sentence in the source or assisting language. This is undesirable as we want the knowledge to be transferred from the parent model (assisting source$\rightarrow$ target) to the child model (source$\rightarrow$target).

In this paper, we propose to pre-order English sentences (assisting language sentences) to match the source language word-order and train the parent model on the pre-ordered corpus. Table \ref{tab:reOrderingRule} shows one of the pre-ordering rules \cite{ananthaGeneric} used along with an example sentence illustrating the effect of pre-ordering. This will ensure that context of words in the parallel source and assisting language sentences are similar, leading to consistent contextual representations across the source languages. Pre-ordering may also be beneficial for other word order divergence scenarios (\textit{e.g., } SOV to SVO), but we leave verification of these additional scenarios for future work.

\section{Experimental Setup}
In this section, we describe the languages experimented with, datasets used, the network hyper-parameters used in our experiments.

\noindent\textbf{Languages}: We experimented with English $\rightarrow$ Hindi translation as the parent task. English is the assisting source language. Bengali, Gujarati, Marathi, Malayalam and Tamil are the source languages, and translation from these to Hindi constitute the child tasks. Hindi, Bengali, Gujarati and Marathi are Indo-Aryan languages, while Malayalam and Tamil are Dravidian languages. All these languages have a canonical SOV word order.

\noindent\textbf{Datasets}: For training English-Hindi NMT systems, we use the IITB English-Hindi parallel corpus \cite{iitbEngHindi} ($1.46M$ sentences from the training set) and the ILCI English-Hindi parallel corpus ($44.7K$ sentences). The ILCI (Indian Language Corpora Initiative) multilingual parallel corpus \cite{jha}\footnote{The corpus is available on request from \url{http://tdil-dc.in/index.php?lang=en}} spans multiple Indian languages from the health and tourism domains. We use the $520$-sentence dev-set of the IITB parallel corpus for validation.  For each child task, we use $2K$ sentences from ILCI corpus as test set.

\noindent\textbf{Network}: We use OpenNMT-Torch \cite{2017opennmt} to train the NMT system. We use the standard encoder-attention-decoder architecture \cite{bahdanau-2014-nmt} with input-feeding approach \cite{DBLP:conf/emnlp/LuongPM15}. The encoder has two layers of bidirectional LSTMs with 500 neurons each and the decoder contains two LSTM layers with 500 neurons each. We use a mini-batch of size $50$ and a dropout layer. We begin with an initial learning rate of $1.0$ and continue training with exponential decay till the learning rate falls below $0.001$. The English input is initialized with pre-trained \textit{fastText} embeddings \cite{grave2018learning} \footnote{\url{https://github.com/facebookresearch/fastText/blob/master/docs/crawl-vectors.md}}. 

English and Hindi vocabularies consists of $0.27M$ and $50K$ tokens appearing at least $2$ and $5$ times in the English and Hindi training corpus respectively. For representing English and other source languages into a common space, we translate each word in the source language into English using a bilingual dictionary (we used \textit{Google Translate} to get single word translations). In an end-to-end solution, it would be ideal to use bilingual embeddings or obtain word-by-word translations \textit{via} bilingual embeddings \cite{2018arXiv180809861X}. However,  publicly available bilingual embeddings for English-Indian languages are not good enough for obtaining good-quality, bilingual representations \cite{smith17b,jawanpuria2018learning} and publicly available bilingual dictionaries have limited coverage. The focus of our study is the influence of word-order divergence on Multilingual NMT. We do not want bilingual embeddings quality or bilingual dictionary coverage to influence the experiments, rendering our conclusions unreliable. Hence, we use the above mentioned large-coverage bilingual dictionary. 

\noindent\textbf{Pre-ordering}: We use  \textit{CFILT-preorder}\footnote{\url{https://github.com/anoopkunchukuttan/cfilt_preorder}} for pre-reordering English sentences. It contains two pre-ordering configurations: (1) \textit{generic} rules (G) that apply to all Indian languages \cite{ananthaGeneric}, and (2) \textit{hindi-tuned} rules (HT) which improves generic rules by incorporating improvements found through error analysis of English-Hindi reordering  \cite{patel2013reordering}. The Hindi-tuned rules improve translation for other English to Indian language pairs too \cite{sataAnuvaadak}.

\begin{table}
\centering
    \resizebox{\columnwidth}{!}{%
    \begin{tabular}{@{}lrrrrrr@{}}
    \toprule
    \multicolumn{1}{c}{\textbf{Language}} & \multicolumn{3}{c}{\textbf{BLEU}} & \multicolumn{3}{c}{\textbf{LeBLEU}} \\ 
    \cmidrule(lr){2-4}
    \cmidrule(lr){5-7}
     & \multicolumn{1}{c}{\multirow{2}{*}{\textbf{\begin{tabular}[c]{@{}c@{}}No\\ Pre-Order\end{tabular}}}} & \multicolumn{2}{c}{\textbf{Pre-Ordered}} &   
     \multicolumn{1}{c}{\multirow{2}{*}{\textbf{\begin{tabular}[c]{@{}c@{}}No\\ Pre-Order\end{tabular}}}} & \multicolumn{2}{c}{\textbf{Pre-Ordered}} \\
     \cmidrule{3-4}\cmidrule{6-7}
     & & \multicolumn{1}{c}{\textbf{HT}} & \multicolumn{1}{c}{\textbf{G}} & & \multicolumn{1}{c}{\textbf{HT}} & \multicolumn{1}{c}{\textbf{G}} \\
    \midrule
    Bengali & 6.72 & 8.83 & \textbf{9.19} & 37.10 & 41.50 & \textbf{42.01} \\
    Gujarati & 9.81 & \textbf{14.34} & 13.90 & 43.21 & 47.36 &  \textbf{47.60} \\
    Marathi & 8.77 & 10.18 & \textbf{10.30} & 40.21 & 41.49 & \textbf{42.22} \\
    Malayalam & 5.73 & 6.49 &  \textbf{6.95} & 33.27 & 33.69 & \textbf{35.09} \\
    Tamil & 4.86 & \textbf{6.04} & 6.00 & 29.38 & 30.77 & \textbf{31.33} \\ \bottomrule
    \end{tabular}%
    }
    \caption{Transfer learning results for \textit{X}-\textit{Hindi} pair, trained on \textit{English}-\textit{Hindi} corpus and sentences from \textit{X} word translated to English.}

    \label{tab:resultsNoDomainDivergence}
\end{table}

\begin{table}
    \centering
    \small
    \begin{tabular}{l r r r}
        \toprule
        \textbf{Language} & \multicolumn{1}{c}{\multirow{2}{*}{\textbf{\begin{tabular}[c]{@{}c@{}}No\\ Pre-Order\end{tabular}}}} & \multicolumn{2}{c}{\textbf{Pre-Ordered}} \\
        \cmidrule{3-4}
        & & \multicolumn{1}{c}{\textbf{HT}} & \multicolumn{1}{c}{\textbf{G}} \\
        \midrule
         Bengali & 1324 & 1139 & 1146 \\
         Gujarati & 1337 & 1190 & 1194 \\
         Marathi & 1414 & 1185 & 1178 \\
         Malayalam & 1251 & 1067 & 1059 \\
         Tamil & 1488 & 1280 & 1252 \\
         \bottomrule
    \end{tabular}
    \caption{Number of UNK tokens generated by each model on the test set.}
    \label{tab:unknownWords}
\end{table}

\begin{figure*}
    \begin{center}
        \includegraphics[trim={4.5cm 9.5cm 1cm 10cm},clip,width=\textwidth,height=\columnwidth]{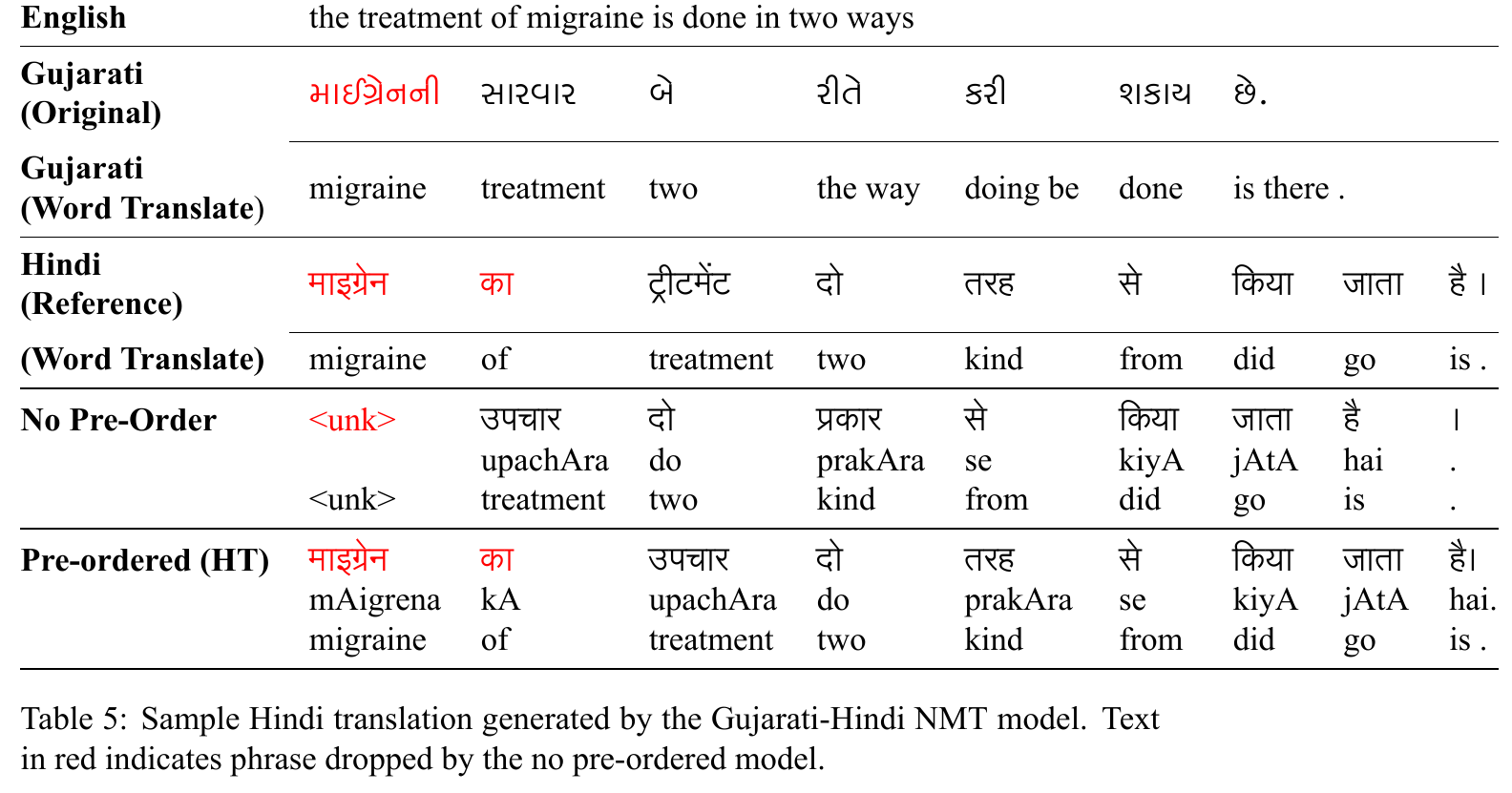}
    \end{center}
    \label{tab:sampleTranslations}
\end{figure*}

\begin{table}[!htb]
    \centering
    \small
    \begin{tabular}{rrrrr}
    \toprule
    \multicolumn{1}{c}{\multirow{2}{*}{\textbf{\begin{tabular}[c]{@{}c@{}}Corpus\\ Size\end{tabular}}}} &  \multicolumn{1}{c}{\multirow{3}{*}{\textbf{\begin{tabular}[c]{@{}c@{}}No \\Transfer\\ Learning\end{tabular}}}} & \multicolumn{1}{c}{\multirow{2}{*}{\textbf{\begin{tabular}[c]{@{}c@{}}No\\ Pre-Order\end{tabular}}}} & \multicolumn{2}{c}{\textbf{Pre-Ordered}}  \\
     \cmidrule{4-5}
     & & & \multicolumn{1}{c}{\textbf{HT}} & \multicolumn{1}{c}{\textbf{G}} \\
     & & & & \\
    \midrule
    \multicolumn{5}{l}{\textbf{\underline{Bengali}}} \\
     - & - & 6.72 & 8.83 & \textbf{9.19} \\
      500 & 0.0 &  11.40 & \textbf{11.49} & 11.00  \\
      1000 & 0.0 &  13.71 & \textbf{13.84} & 13.62  \\
      2000 & 0.0 &  16.41 & \textbf{16.79} & 16.01  \\
      3000 & 0.0 &  17.44 & \textbf{18.42}$\dagger$ & 17.82  \\
      4000 & 0.0 &  18.86 & \textbf{19.17} & 18.66  \\
      5000 & 0.07 &  19.58 & \textbf{20.15}$\dagger$ & 19.82  \\
      10000 & 1.87 &  22.50 & \textbf{22.92} & 22.53  \\
     \midrule
     \multicolumn{5}{l}{\textbf{\underline{Gujarati}}} \\
     - & - & 9.81 & \textbf{14.34} & 13.90 \\
       500 & 0.0 & 17.27 & 17.11 & \textbf{17.75} \\
      1000 & 0.0 & 21.68 & \textbf{22.12} & 21.45 \\
      2000 & 0.0 & 25.34 & \textbf{25.73} & 25.63 \\
      3000 & 0.29 & 27.48 & 27.77 & \textbf{27.83} \\
      4000 & 0.82 & 29.20 & 29.49 & \textbf{29.51} \\
      5000 & 0.0 & 29.87 & \textbf{31.09}$\dagger$ & 30.58$\dagger$ \\
      10000 & 1.52 & 33.97 & \textbf{34.25} & 34.08 \\
     \midrule
     \multicolumn{5}{l}{\textbf{\underline{Marathi}}} \\
     - & - & 8.77 & 10.18 & \textbf{10.30} \\
       500 & 0.0 & 12.84 & \textbf{13.61}$\dagger$ & 12.97 \\
      1000 & 0.0 & 15.62 & 15.75 & \textbf{16.10}$\dagger$ \\
      2000 & 0.0 & 18.59 & \textbf{19.10} & 18.67 \\
      3000 & 0.0 & 20.51 & \textbf{20.76} & 20.29 \\
      4000 & 0.24 & \textbf{21.78} & 21.77 & 21.39 \\
      5000 & 0.29 & 22.21 & 22.41 & \textbf{22.73}$\dagger$ \\
      10000 & 7.90 & 25.16 & \textbf{25.88} & 25.36  \\
      \midrule
     \multicolumn{5}{l}{\textbf{\underline{Malayalam}}} \\
     - & - & 5.73 & 6.49 & \textbf{6.95} \\
       500 & 0.0 & 5.40 & 5.54 & \textbf{6.17}$\dagger$ \\
      1000 & 0.0 & 7.34 & 7.36 & \textbf{7.63} \\
      2000 & 0.0 & 8.24 & \textbf{8.66}$\dagger$ & 8.31  \\
      3000 & 0.0 & 9.11 & 9.30 & \textbf{9.31}  \\
      4000 & 0.0 & 9.65 & \textbf{9.91} & 9.87  \\
      5000 & 0.03 & 10.26 & \textbf{10.47} & 10.28  \\
      10000 & 0.0 & \textbf{11.96} & 11.85 & 11.63  \\
      \midrule
     \multicolumn{5}{l}{\textbf{\underline{Tamil}}} \\
     - & - & 4.86 & \textbf{6.04} & 6.00 \\
       500 & 0.0 & 5.49 & \textbf{5.85}$\dagger$ & 5.59 \\
      1000 & 0.0 & 7.04 & 7.23 & \textbf{7.44}$\dagger$ \\
      2000 & 0.0 & 8.83 & 8.84 & \textbf{9.24} \\
      3000 & 0.0 & 9.80 & \textbf{10.04} & 9.56 \\
      4000 & 0.0 & 9.69 & \textbf{10.59}$\dagger$ & 10.25$\dagger$ \\
      5000 & 0.03 & 10.84 & \textbf{10.93} & 10.69 \\
      10000 & 0.0 & 12.71 & \textbf{13.05} & 12.69  \\
    \bottomrule
    \end{tabular}
    \caption{Transfer learning results (BLEU) for \textit{Indian Language}-\textit{Hindi} pair, fine-tuned with varying number of \textit{Indian Language}-\textit{Hindi} parallel sentences. $\dagger$Indicates statistically significant difference between \textit{Pre-ordered} and \textit{No Pre-ordered} results using paired bootstrap resampling \cite{koehnStatTest} for a $p$-value less than $0.05$. \textit{No Transfer Learning} model refers to training the model on varying number of \textit{Indian Language}-\textit{Hindi} parallel sentences with randomly initialized weights.}
    \label{tab:bengaliResults}
\end{table}

\section{Results}
We experiment with two scenarios: (a) an extremely resource scarce scenario with no parallel corpus for child tasks, (b) varying amounts of parallel corpora  available for child task.

\subsection{No Parallel Corpus for Child Task}
The results from our experiments are presented in the Table \ref{tab:resultsNoDomainDivergence}. We report BLEU scores and LeBLEU\footnote{LeBLEU (Levenshtein Edit BLEU) is a variant of BLEU that does a soft-match of reference and output words based on edit distance, hence it can handle morphological variations and cognates \cite{virpioja2015lebleu}.} scores. We observe that both the pre-ordering models significantly improve the translation quality over the no-preordering models for all the language pairs. The results support our hypothesis that word-order divergence can limit the benefits of multilingual translation. Thus, reducing the word order divergence improves translation in extremely low-resource scenarios. 

An analysis of the outputs revealed that pre-ordering significantly reduced the number of UNK tokens (placeholder for unknown words) in the test output (Table \ref{tab:unknownWords}). We hypothesize that due to word order divergence between English and Indian languages, the encoder representation generated is not consistent leading to decoder generating unknown words. However, the pre-ordered models generate better encoder representations leading to lesser number of UNK tokens and better translation, which is also reflected in the BLEU scores and Table \ref{tab:sampleTranslations}.

\subsection{Parallel Corpus for Child Task}
We study the impact of child task parallel corpus on pre-ordering. To this end, we fine-tune the parent task model with the child task parallel corpus. Table \ref{tab:bengaliResults} shows the results for \textit{Bengali-Hindi}, \textit{Gujarati-Hindi}, \textit{Marathi-Hindi}, \textit{Malayalam-Hindi}, and \textit{Tamil-Hindi} translation. 
We observe that pre-ordering is beneficial when almost no child task corpus is available. As the child task corpus increases, the model learns the word order of the source language; hence, the non pre-ordering models perform almost as good as or sometimes better than the pre-ordered ones. The non pre-ordering model is able to forget the word-order of English and learn the word order of Indian languages.
We attribute this behavior of the non pre-ordered model to the phenomenon of catastrophic forgetting \cite{MCCLOSKEY1989109,FRENCH1999128} which enables the model to learn the word-order of the source language when sufficient child task parallel corpus is available.

We also compare the performance of the fine-tuned model with the model trained only on the available source-target parallel corpus with randomly initialized weights (No Transfer Learning). Transfer learning, with and without pre-ordering, is better compared to training only on the small source-target parallel corpus.

\section{Conclusion}
In this paper, we show that handling word-order divergence between the source and assisting languages is crucial for the success of multilingual NMT in an extremely low-resource setting. We show that pre-ordering the assisting language to match the word order of the source language significantly improves translation quality in an extremely low-resource setting. 
If pre-ordering is not possible, fine-tuning on a small source-target parallel corpus is sufficient to overcome word order divergence. 
While the current work focused on Indian languages, we would like to validate the hypothesis on a more diverse set of languages. We would also like to explore alternative methods to address word-order divergence which do not require expensive parsing of the assisting language corpus. Further, use of pre-ordering to address  word-order divergence for multilingual training of other NLP tasks can be explored. 

\section*{Acknowledgements}
We would like to thank Raj Dabre for his helpful suggestions and comments.

\bibliography{naaclhlt2019}
\bibliographystyle{acl_natbib}

\end{document}